\documentclass[11pt,a4paper]{article}

\usepackage{mathtools,cancel}

\usepackage{soul,xcolor}
\usepackage{xcolor}
\usepackage{multirow}
\usepackage{pstricks}
\usepackage{dcolumn}
\usepackage{bm}
\usepackage{latexsym}
\usepackage{dcolumn}
\usepackage[utf8x]{inputenc} 
\usepackage{amsmath}
\usepackage{amsfonts,amssymb}
\usepackage{graphicx,epsfig}
\usepackage{color}
\usepackage{psfrag}
\usepackage{amsthm}
\usepackage{ulem} 
\usepackage{soul} 
\usepackage{flushend}
\usepackage{titlesec}
\usepackage{slashed}
\usepackage{authblk}
\usepackage[font=footnotesize]{caption}
\usepackage[hmarginratio=1:1,top=32mm,columnsep=60pt]{geometry}

\newcommand{\bea}{\begin{eqnarray}}
\newcommand{\eea}{\end{eqnarray}}
\newcommand{\be}{\begin{equation}}
\newcommand{\ee}{\end{equation}}
\newcommand{\ba}{\begin{array}}
\newcommand{\ea}{\end{array}}
\usepackage{comment}
\usepackage{array}
\usepackage{makecell}
\usepackage{enumitem}

\def\beq{\begin{equation}}
\def\eeq{\end{equation}}
\def\bea{\begin{eqnarray}}
\def\eea{\end{eqnarray}}

\begin{document}

\unitlength = 1mm

\title{\textbf{Analyzing Fairness of Computer Vision and Natural Language Processing Models}}

\author{\normalsize Ahmed Rashed$^a$\footnote{amrashed@ship.ed}, Abdelkrim Kallich$^a$\footnote{ak2206@ship.edu}, and Mohamed Eltayeb$^{b,c}$\footnote{443059243@stu.iu.edu.sa  }}
\affil{\small
$^a$Department of Physics, Shippensburg University of Pennsylvania,\\
Franklin Science Center, 1871 Old Main Drive, Pennsylvania, 17257, USA\\
$^b$Islamic University of Madinah, Medina, Al Jamiah, Madinah 42351, Saudi Arabia\\
$^c$University of Khartoum, Khartoum, Al-Nil Avenue, Khartoum 11115, Sudan

 }
\date{}

{\let\newpage\relax\maketitle}

\begin{abstract}
\noindent \normalsize

Machine learning (ML) algorithms play a critical role in decision-making across
various domains, such as healthcare, finance, education, and law enforcement. However,
concerns about fairness and bias in these systems have raised significant ethical and social
challenges. To address these challenges, this research utilizes two prominent fairness
libraries, Fairlearn by Microsoft and AIF360 by IBM. These libraries offer comprehensive
frameworks for fairness analysis, providing tools to evaluate fairness metrics, visualize
results, and implement bias mitigation algorithms. The study focuses on assessing and mitigating
biases for unstructured datasets using Computer Vision (CV) and Natural Language
Processing (NLP) models. The primary objective is to present a comparative analysis of the
performance of mitigation algorithms from the two fairness libraries. This analysis involves
applying the algorithms individually, one at a time, in one of the stages of the ML lifecycle,
pre-processing, in-processing, or post-processing, as well as sequentially across more than
one stage. The results reveal that some sequential applications improve the performance of
mitigation algorithms by effectively reducing bias while maintaining the model’s performance.
Publicly available datasets from Kaggle were chosen for this research, providing a
practical context for evaluating fairness in real-world machine learning workflows.

\vspace{1 cm}




\end{abstract}


\newpage

\section{Introduction}
\label{sec:introduction}

Machine learning (ML) algorithms are extensively employed across various fields, including entertainment, shopping, healthcare, finance, education, and law enforcement, as well as in critical areas such as loan approvals [1] and hiring decisions [2, 3]. They offer benefits like consistent performance and the ability to analyze numerous variables [4, 5]. However, these systems are not immune to bias, which can lead to unfair outcomes [6, 7]. Bias in ML is particularly concerning when decisions directly affect individuals or communities, potentially resulting in discrimination. Ensuring that these systems operate ethically and equitably is crucial. Fair decision-making necessitates impartiality, avoiding favoritism based on inherent or acquired characteristics. Biased algorithms deviate from this principle, skewing decisions in favor of certain groups.

The idea of "fairness" in algorithmic systems is deeply tied to the sociotechnical context. Different forms of fairness-related harms have been identified in Table 1.
\begin{table}[h!]
\centering
\begin{tabular}{|p{3cm}|p{10cm}|}
\hline
\textbf{Harm Type} & \textbf{Definition} \\ \hline
\textbf{Allocation Harm} & Unequal distribution of opportunities or resources, such as an algorithm disproportionately favoring men over women for job offers [8]. \\ \hline
\textbf{Quality-of-Service Harm} & Unequal performance across groups, e.g., facial recognition systems misclassifying Black women more frequently than White men [9], or speech recognition systems underperforming for users with speech disabilities [10]. \\ \hline
\textbf{Stereotyping Harm} & Reinforcing societal stereotypes, such as image searches for "CEO" showing predominantly White men [8]. \\ \hline
\textbf{Denigration Harm} & Producing offensive outputs, such as misclassifying individuals as animals or chatbots generating slurs [8]. \\ \hline
\textbf{Representation Harm} & Over- or under-representing certain groups, such as racial bias in welfare fraud investigations or neglecting elderly populations in surveillance applications [8]. \\ \hline
\textbf{Procedural Harm} & Violations of social norms in decision-making practices, such as penalizing job candidates for extensive experience or failing to ensure transparency and accountability in algorithmic decisions [11]. \\ \hline
\end{tabular}
\caption{Types of Harm and Their Definitions}
\label{tab:harm_definitions}
\end{table}
These harms often overlap and highlight the importance of addressing fairness concerns from the development phase of ML systems.

Incorporating ML fairness techniques as an industry application in a research paper is essential for advancing ethical AI. Fairness libraries provide tools to evaluate and mitigate biases in ML models, a critical need as industries increasingly rely on AI-driven decision-making. These tools are particularly valuable for ensuring equity in sectors such as finance, banking, and healthcare. They offer intuitive and interactive ways to explore model behavior, enabling stakeholders to assess fairness trade-offs effectively. By demonstrating the practical applications of these tools, researchers can bridge the gap between academic innovation and industry adoption, fostering the development of transparent and fair AI systems. This approach not only underscores the versatility of fairness libraries but also highlights their importance in addressing real-world fairness challenges across industries.

Therefore, in this paper, we present use cases of three fairness libraries—Fairlearn (Microsoft), AIF360 (IBM), and What-If Tool (Google)—which are trusted in the industry for assessing and mitigating bias in ML models before deployment. This work aims to encourage industry professionals to implement these libraries across various applications.

In our study, we conducted a comparative evaluation of two approaches to bias mitigation in ML models. Specifically, we analyzed the effectiveness of applying a single mitigation algorithm at a time versus the sequential application of multiple mitigation algorithms across different stages of the ML lifecycle: pre-processing, in-processing, and post-processing. The sequential approach is designed to harness the strengths of stage-specific algorithms, enabling a more comprehensive bias mitigation strategy.

Figure 1 illustrates the ML lifecycle and the integration of pre-processing, in-processing, and post-processing mitigation algorithms. It depicts where a mitigation algorithm can be applied within a single ML stage (individual application) and how multiple algorithms can be implemented across different stages simultaneously (sequential application).

\begin{figure}[!h]
\centering
\includegraphics[scale=0.4]{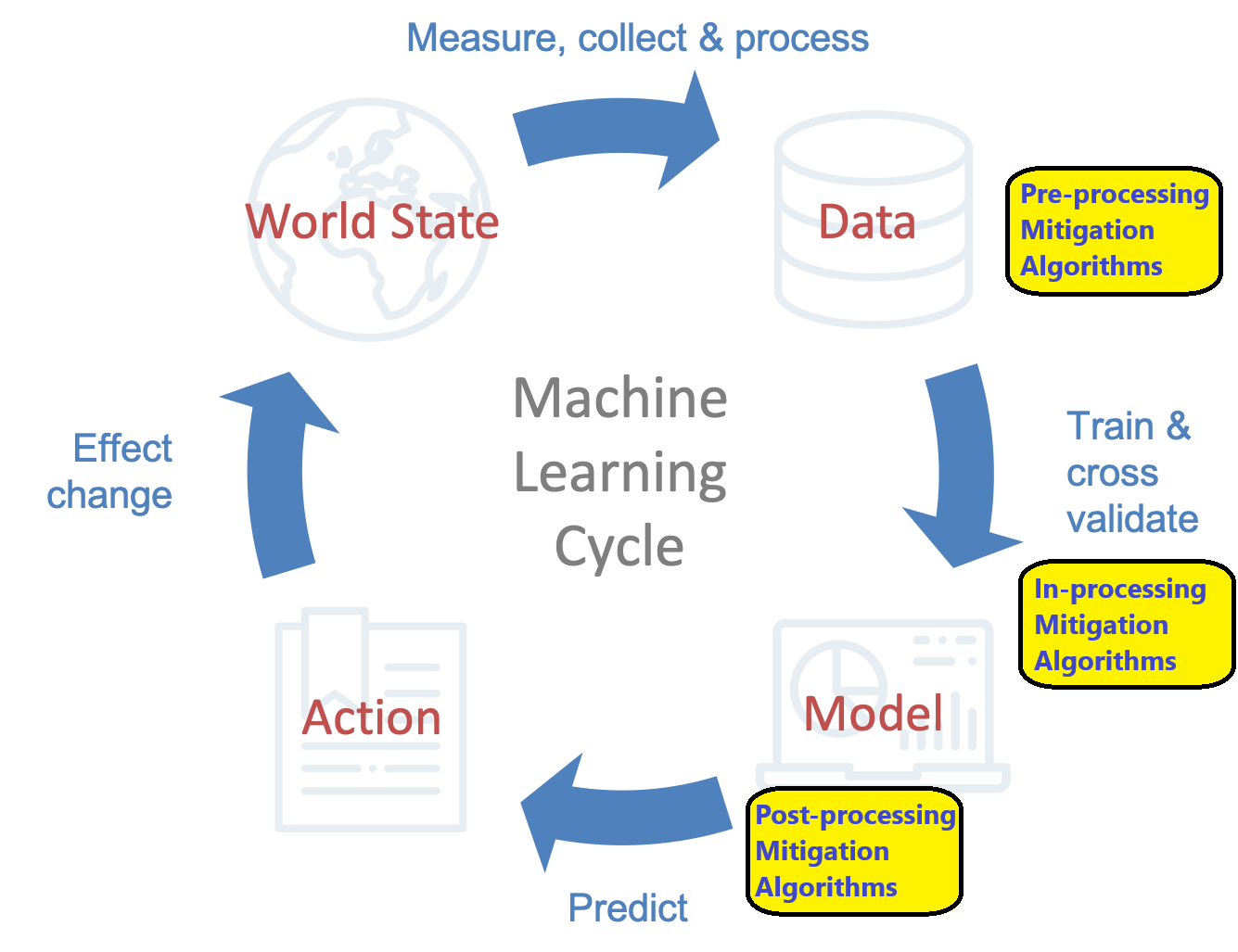}
\caption{The ML lifecycle and the application of the pre-processing, in-processing, and post-processing mitigation algorithms.}
\end{figure}

While many previous studies have focused on applying fairness interventions at individual stages of the ML lifecycle, our study advances the field by implementing these interventions sequentially across three stages. This comprehensive framework significantly enhances bias mitigation by addressing biases holistically, reducing the risk of fairness issues propagating through the model development process, and minimizing residual disparities that may persist with single-stage interventions [12, 13]. Applying this lifecycle-based approach to real-world datasets provides strong empirical evidence of its effectiveness. Our work bridges the gap between theoretical fairness concepts and practical implementation, encouraging broader adoption of lifecycle-based fairness practices [14].
The objectives of this study are:
\begin{enumerate}
\item The central objective is to demonstrate how a sequential application of fairness mitigation algorithms across the ML lifecycle stages, pre-processing, in-processing, and post-processing, sometimes leads to good mitigation of biases compared to applying these methods in isolation. 
\item	To measure the level of bias present in ML models trained on the chosen dataset.
\item	To compare the performance and effectiveness of the three fairness libraries in identifying and mitigating bias.
\item	To offer actionable insights on implementing fairness tools in real-world ML workflows.
\end{enumerate}

This study examines the fairness of ML models applied to unstructured datasets in computer vision (CV) and natural language processing (NLP) tasks. Unstructured datasets, known for their lack of organization, facilitate efficient analysis and are commonly used in AI applications. However, these datasets often reflect historical biases or systemic discrimination, which can impact the fairness of AI models, making robust evaluation and mitigation essential. A similar fairness analysis for structured datasets was performed in our previous work [15].

For this research, a publicly available Kaggle dataset [16] was selected. Kaggle datasets provide realistic scenarios for evaluating ML models and are well-suited for fairness studies. The dataset underwent preprocessing and was used to develop NLP and CV models. NLP and CV tasks are particularly significant in fairness research, as biased predictions can disproportionately affect specific groups.

To identify and address biases, the study employed three advanced fairness libraries: Fairlearn by Microsoft, AIF360 by IBM, and the What-If Tool by Google. These libraries offer tools for assessing fairness metrics, visualizing model behavior, and applying bias mitigation techniques. By leveraging these tools, the research systematically evaluates the fairness of NLP and CV models and investigates strategies to minimize bias.

The remainder of this paper is structured as follows: Section 2 reviews related work on ML fairness. Section 3 details the methodology, including dataset preprocessing and model development. Section 4 discusses fairness analyses using the selected libraries and their functionalities. Section 5 presents a comparative evaluation of the libraries and their results. Future work is discussed in Section 6. Finally, Section 7 concludes with a summary of the findings.

\section{Review of Related Work}

Bias in ML models has become an increasingly urgent concern, particularly as these models impact critical areas such as healthcare, hiring, and criminal justice. A growing body of research has explored the origins, manifestations, and mitigation strategies for bias, providing a solid foundation for addressing this widespread challenge [17].

One key area of research focuses on identifying and categorizing biases in ML models. Reference [18] introduces a taxonomy of biases, classifying them into historical, representation, and measurement biases. Historical bias arises from pre-existing inequities in the data, even before the application of ML techniques. Representation bias occurs when certain groups are under- or overrepresented in the training data, leading to skewed predictions [19]. Measurement bias results from inaccuracies in the features or labels used for training, often caused by flawed measurement processes.

Another line of research explores methods for detecting bias. Tools such as disparate impact analysis [20] and fairness metrics like demographic parity, equal opportunity, and disparate mistreatment [21] are commonly used. In structured datasets, researchers focus on two fairness dimensions: group fairness, which ensures equitable treatment across demographic groups, and individual fairness, which treats similar individuals similarly [22]. Ref. [23] examines the conflicts between these fairness definitions, emphasizing the need for context-specific trade-offs.

The technical challenges of mitigating bias have also been extensively studied. Pre-processing techniques, such as re-weighting data samples or modifying labels to enhance fairness, address biases at the dataset level [24]. In-processing methods, such as adversarial debiasing [25], incorporate fairness constraints directly into model training. Post-processing approaches modify model outputs to meet fairness criteria, including re-ranking methods proposed by [21]. However, Ref. [26] highlights the inherent trade-offs between fairness and accuracy, demonstrating the challenge of optimizing both simultaneously.

Bias analysis in structured datasets has received significant attention due to the prevalence of tabular data in decision-making systems. These datasets often carry latent biases reflecting historical inequities or systemic discrimination. For example, the COMPAS dataset, used in criminal justice, exhibits racial disparities in predictive outcomes [27]. Research has also examined how feature selection and preprocessing can either mitigate or exacerbate biases. Ref. [28] explores the impact of feature correlation with sensitive attributes on fairness and proposes methods to disentangle these relationships.

Recent advances have emphasized interpretability as a critical component of bias analysis. Ref. [29] introduced LIME (Local Interpretable Model-agnostic Explanations), which helps stakeholders understand and detect biased patterns in model predictions. Similarly, Ref. [30] developed SHAP (SHapley Additive exPlanations), providing consistent and accurate feature importance values. These interpretability tools have been instrumental in identifying biases in structured datasets by allowing detailed analysis of how features contribute to unfair outcomes.

Multiple demographic factors have also emerged as a key consideration in bias studies. Ref. [31] emphasizes the necessity of evaluating models across various demographic dimensions, revealing compounded disparities for groups such as Black women in facial recognition systems. For structured datasets, Ref. [32] proposes fairness-enhancing strategies that account for multiple subgroups simultaneously, overcoming the limitations of single-axis fairness analyses.

The literature on bias in ML covers a wide range of topics, including foundational concepts, detection methods, mitigation strategies, and transparency tools. Despite significant progress, challenges remain in applying these techniques to structured datasets, particularly in balancing fairness with competing objectives like accuracy and transparency. This review underscores the ongoing need for research into comprehensive, context-aware methods for understanding and mitigating bias in ML models.

\section{Dataset and Model Details}
\label{sec:meth}

For the computer vision model, we used the UTKFace dataset [16], a collection of over 20,000 facial images labeled with demographic information, including age, gender, and ethnicity. Widely used for ML tasks such as facial analysis and demographic prediction, the dataset includes images cropped and aligned to ensure consistent analysis across models.

This dataset is commonly used for tasks like age and gender prediction, facial recognition, and demographic analysis. Researchers often apply deep learning techniques, including convolutional neural networks (CNNs), to extract features and train models for these predictive tasks.

For the Natural Language Processing (NLP) model, we used the California Independent Medical Review (IMR) dataset [16], available on Kaggle. This dataset contains healthcare-related data focused on medical review decisions, specifically text-based reviews related to insurance appeals. It is particularly valuable for text classification, sentiment analysis, and decision-making predictions in a healthcare context.

Although the dataset consists of text entries, researchers often use ML models such as Logistic Regression, Random Forest, or advanced NLP models like BERT to analyze decision outcomes and identify patterns in the appeals based on textual data.

For the computer vision use case, we selected the UTKFace dataset because of its broad demographic coverage. Labeled with age, gender, and ethnicity, it is especially suitable for analyzing biases in facial recognition models. Its diversity allows for a comprehensive evaluation of fairness issues, particularly regarding demographic disparities. By employing state-of-the-art facial recognition models, we aim to examine their performance across different demographic groups and explore how fairness interventions can address any observed disparities.

For the NLP use case, we chose the California Independent Medical Review dataset because it provides a rich text corpus of medical reviews, which is critical for assessing fairness in healthcare-related NLP applications. This dataset represents sensitive contexts where biased model behavior could significantly impact decision-making processes, especially in healthcare, which affects diverse patient populations. We pair this dataset with widely used NLP models to assess how fairness techniques can mitigate biases that may disproportionately impact specific groups.

We chose these datasets because they are highly relevant to real-world applications where fairness is crucial. The UTKFace dataset represents potential fairness challenges in public or commercial systems, such as those used in security or customer service. Similarly, the California Independent Medical Review dataset highlights fairness issues in healthcare, where equitable outcomes are essential. By focusing on these datasets and domains, our research aims to offer actionable insights that can be applied to similar scenarios in industry.

\section{Implementation of Fairness Analyses}

In recent years, the growing use of ML across various sectors has raised significant concerns regarding fairness and bias in classification models. Given their potential impact on decision-making in critical areas such as healthcare, finance, and criminal justice [33], ensuring fairness in these models is crucial. Bias in ML models can arise from factors such as imbalanced training data, model selection, and inherent societal biases, leading to discriminatory outcomes against marginalized groups [34].

In this study, we conduct a comparative evaluation to investigate the effectiveness of different strategies for mitigating bias in ML models. The methodology involves two distinct approaches:

\begin{enumerate}
\item \textbf{Single-Algorithm Application:} We apply individual mitigation algorithms independently at each stage of the ML lifecycle—pre-processing, in-processing, or post-processing. This approach allows us to evaluate the impact of each algorithm in isolation.

\item \textbf{Sequential Algorithm Application:} We implement a sequential approach by applying multiple mitigation algorithms across the ML lifecycle stages. Specifically, we combine pre-processing, in-processing, and post-processing techniques in a systematic sequence to leverage the unique strengths of each algorithm.
\end{enumerate}

For both approaches, we evaluate their performance using standard fairness metrics and monitor their impact on model accuracy. The sequential strategy aims to address bias more comprehensively by integrating mitigation efforts across different stages, ensuring a holistic reduction of bias while maintaining model performance.

This methodology provides a robust framework to compare the efficacy of standalone and sequential applications of fairness interventions, offering actionable insights for developing equitable ML systems.

To address these challenges, several libraries and tools have been developed to help practitioners analyze and reduce bias in their models. Notable among them are Fairlearn, AIF360, and the What-If Tool, each offering unique features for evaluating and enhancing fairness. Details on the libraries, datasets, metrics used, and mitigation algorithms employed are provided in Table \ref{tab:harm_definitions}.
\begin{table}[h!]
\centering
\begin{tabular}{|>{\raggedright\arraybackslash}p{3.3cm}|>{\raggedright\arraybackslash}p{3.3cm}|>{\raggedright\arraybackslash}p{3.3cm}|>{\raggedright\arraybackslash}p{3.3cm}|}
\hline
\textbf{Library} & \textbf{Dataset} & \textbf{Metrics} & \textbf{Mitigation Algorithms} \\ \hline
\textbf{Fairlearn} by Microsoft [35] & CV: UTKFace dataset [16]. ~~~~ NLP: California Independent Medical Review (IMR) Dataset [16] & Demographic Parity, Equalized Odds & Correlation Remover, Exponentiated Gradient, Grid Search, and Threshold Optimizer. \\ \hline
\textbf{AIF360} by IBM [36] & CV: UTKFace dataset [16]. ~~~~ NLP: California Independent Medical Review (IMR) Dataset [16] & Statistical Parity Difference, Average Odds Difference, Equal Opportunity Difference, Theil Index, and Generalized Entropy Index. & Reweighing, Disparate Impact Remover, Learning Fair Representations, Adversarial Debiasing, Equalized Odds, and Reject Option Classification. \\ \hline
\textbf{What-If Tool} by Google [37] & CV: UTKFace dataset [16]. ~~~~ NLP: California Independent Medical Review (IMR) Dataset [16] & Demographic Parity Difference, Equalized Odds Difference & N/A \\ \hline
\end{tabular}
\caption{Fairness Analysis Tools}
\label{tab:harm_definitions}
\end{table}

These tools are essential for researchers seeking to identify and mitigate bias in ML classification models, providing the methodologies needed to ensure equitable AI systems and informed decision-making. Table 3 summarizes the strengths and weaknesses of the libraries used in this analysis: Fairlearn and AIF360, along with the What-If Tool by Google. While the What-If Tool is valuable for model evaluation, it is limited in terms of mitigation algorithms, making it less suitable for this study.

\begin{table}[h!]
\begin{tabular}{|>{\raggedright\arraybackslash}p{0.7cm}|>{\raggedright\arraybackslash}p{4cm}|>{\raggedright\arraybackslash}p{4cm}|>{\raggedright\arraybackslash}p{4cm}|>{\raggedright\arraybackslash}p{4cm}|}
\hline
 & \textbf{Fairlearn} & \textbf{AIF360} &  \textbf{What-If-Tool} \\ \hline
Pro   & \begin{itemize}[left=0pt] \item Easy to use. Useful for educational purpose \end{itemize} & \begin{itemize}[left=0pt] \item Large number of metrics and mitigation algorithms. \item Can handle all classification tasks.  \end{itemize} & \begin{itemize}[left=0pt] \item Interactive tool with visualizations for model's predictions. \end{itemize} \\ \hline
Cons  & \begin{itemize}[left=0pt] \item Limited metrics and mitigation algorithms. \item Very slow with big datasets \item SUpports only binary classification tasks. \end{itemize} & \begin{itemize}[left=0pt] \item Some algorithms needs high computational resources (e.g. Learning Fair Representation) \end{itemize} & \begin{itemize}[left=0pt] \item Limited in terms of mitigation algorithms. \item Outdated library. \end{itemize} \\ \hline
\end{tabular}
\caption{Results of Fairlearn + Computer Vision with applying the mitigation algorithms one at a time.}
\label{tab:harm_definitions1asdasd}
\end{table}

\section{Results and Discussions}

This section presents the results of applying fairness libraries to computer vision (CV) and natural language processing (NLP) models. The primary objective of this analysis is to improve fairness metrics while maintaining or enhancing performance metrics. Detailed code and implementation specifics are provided in [38].

For the CV model, the sensitive features analyzed for potential bias are gender and ethnicity, with the target variable being the prediction of a person’s age group. In the NLP model, the target variable is determining whether a patient requires urgent intervention, based on a combination of the doctor’s findings, patient gender, and age group. The sensitive features here are gender and age group, with the goal of identifying and addressing any model bias related to these attributes.

The following discussion highlights the results obtained from applying fairness libraries to both the CV and NLP models.

For \textbf{Fairlearn}, accuracy was used to evaluate the overall predictive performance, while demographic parity difference served as a fairness metric to measure disparities across demographic groups. Mitigation strategies were implemented at three stages of the ML pipeline: preprocessing, in-processing, and post-processing:
\begin{itemize}
\item	\textbf{Preprocessing}: Techniques like the Correlation Remover addressed biases by eliminating correlations between sensitive and non-sensitive features while preserving data integrity [39].
\item	\textbf{In-processing}: Algorithms such as Exponentiated Gradient incorporated fairness constraints during model training to ensure fairer decision boundaries [40].
\item	\textbf{Postprocessing}: Methods like the Threshold Optimizer adjusted predictions after model training to satisfy fairness criteria precisely, ensuring no residual disparities [41].
\end{itemize}
The evaluation identified the most effective algorithms as those balancing high predictive accuracy with reduced bias, demonstrating the importance of an integrated fairness approach to mitigate biases in structured datasets.

For \textbf{AIF360}, mitigation algorithms were applied at multiple stages to address bias comprehensively:
\begin{itemize}
\item	\textbf{Reweighing}: It assigns weights to training instances to correct imbalances in demographic representation, addressing biases embedded in the dataset.
\item	\textbf{Equalized Odds}: It imposes constraints during model training to align true positive and false positive rates across sensitive groups, enhancing fairness without significantly reducing accuracy.
\item	\textbf{Disparate Impact Ratio}: Ratio of selection rates  which is the percentage of samples with positive selection.
\item	\textbf{Learning Fair Representations (Consistency Score)}: The metric computes the consistency score. Individual fairness metric that measures how similar the (predicted) labels are for similar instances (records). It compares a model’s classification prediction of a given data item x to its k-nearest neighbors, kNN(x). It applies the kNN function to the full set of examples to obtain the most accurate estimate of each point’s nearest neighbors
\item	\textbf{Adversarial Debiasing}: Adversarial debiasing is an in-processing technique that utilizes adversarial learning to mitigate bias. The method involves training a predictive model (the classifier) while simultaneously training an adversarial model that seeks to predict the sensitive attribute (e.g., gender or race) from the classifier’s predictions. The classifier is then optimized to minimize its primary objective (e.g., accuracy) while also minimizing the adversary's ability to detect the sensitive attribute. This process reduces the dependency of the classifier's predictions on sensitive features, thereby enhancing fairness.
\item 	\textbf{Equalized Odds}: Equalized odds is a fairness constraint that ensures the predictive performance of a model is consistent across all groups defined by a sensitive attribute. Specifically, it requires that the true positive rate (TPR) and false positive rate (FPR) are equal for all groups. By enforcing parity in these rates, equalized odds minimizes discrimination and ensures that the model's predictive behavior is fair across different demographic groups.
\item	\textbf{Reject Option Classification (ROC)}: Reject option classification is a postprocessing technique that adjusts the decision boundary of a trained classifier. It introduces a "reject option" for samples with a high likelihood of misclassification or bias. In this option, predictions for individuals from an unprivileged group with borderline scores are reclassified to ensure fairness. ROC aims to correct unfair outcomes by altering decisions in favor of fairness, often at the cost of slight performance reductions.
\end{itemize}

\subsection{Bias Mitigation in Computer Vision Models}
\subsubsection{Fairlearn Library}
The Fairlearn library was used to evaluate fairness in a computer vision model designed for age group classification. The dataset consisted of four age groups: 0–18, 18–30, 30–80, and above 80 years. Since Fairlearn does not support multi-class classification directly, metrics were computed separately for each age group. In Figure 2, we present the values of the demographic parity and equalized odds metrics for each subgroup of the sensitive feature (age group). The upper panel displays the demographic parity values for each subgroup, starting with the original model (accuracy = 0.66). We then show the metrics after applying different mitigation algorithms, such as Correlation Remover, Exponentiated Gradient, Grid Search, and Threshold Optimizer. The lower panel shows the equalized odds values for the same cases. Figure 3 presents the effect of applying the mitigation algorithms in sequential order. The results are discussed below.

\begin{enumerate}
\item	Metrics:
\begin{itemize}
\item	Demographic Parity Difference (DPD): This measures the difference in selection rates among groups within a sensitive feature, such as age group. A small DPD indicates reduced bias, meaning the model treats all groups equally.
\item	Equalized Odds Difference (EOD): This evaluates the model's performance, comparing false positive and true positive rates across groups within a sensitive feature.
\end{itemize}
\item	Results:\\
The fairness metrics, applying the mitigation algorithms both individually and sequentially, are shown in Figures 2 and 3, with a summary of the results in Tables (\ref{tab:harm_definitions1}, \ref{tab:harm_definitions2}).
\item	Insights:\\
From the results in Tables (\ref{tab:harm_definitions1}, \ref{tab:harm_definitions2}), we find that the original model has accuracy of 66\%.

\begin{itemize}
\item Original Model: Accuracy = 66\%.
  
\item Demographic Parity Difference (DPD): 
\begin{itemize}
 \item The highest DPD was 55\%.
  \item After applying the mitigation algorithms individually, the highest DPD decreased to 3.5\% (best case scenario) with the Threshold Optimizer algorithm, but accuracy dropped to 18\%.
  \item When applying the mitigation algorithms sequentially, the highest DPD decreased to 5\% (best case scenario) with the combination of Correlation Remover + Exponentiated Gradient, while accuracy increased to 30.1\%.
  \item Comparing the individual and sequential approaches: The individual application reduced bias by 93.63\%, but accuracy dropped by 72.72\%. The sequential application reduced bias by 90.91\%, with a smaller accuracy decrease of 54.39\%. This suggests that the sequential approach maintains model performance better than the individual approach while achieving similar levels of bias reduction.
\end{itemize}

\item Equalized Odds Difference (EOD):
\begin{itemize} 
  \item The highest EOD was 31.1\%.
  \item After applying the mitigation algorithms individually, the highest EOD decreased to 3.4\% (best case scenario) with the Threshold Optimizer algorithm, but accuracy dropped to 18\%.
  \item When applying the mitigation algorithms sequentially, the highest EOD decreased to 7.8\% (best case scenario) with Correlation Remover + Exponentiated Gradient, and accuracy increased to 30.1\%.
  \item Comparing the individual and sequential approaches: The individual application reduced bias by 89.07\%, but accuracy dropped by 72.72\%. The sequential application reduced bias by 74.92\%, with a smaller accuracy decrease of 54.39\%. This shows that while the sequential approach maintains performance better, it does not reduce bias as effectively as the individual approach.
\end{itemize}
\end{itemize}

\end{enumerate}

\begin{table}[h!]
\hspace{-1cm}
\begin{tabular}{|>{\raggedright\arraybackslash}p{3cm}|>{\raggedright\arraybackslash}p{2cm}|>{\raggedright\arraybackslash}p{2cm}|>{\raggedright\arraybackslash}p{3.5cm}|>{\raggedright\arraybackslash}p{3.5cm}|}
\hline
\textbf{Metric} & \textbf{Accuracy of Original Model} & \textbf{Subgroup with Max Bias} &  \textbf{Algorithm with Max Mitigation} & \textbf{Algorithm with Min Mitigation} \\ \hline
Demographic Parity Difference (DPD) (\%) & 66\% &   55\% (30-80y) & Threshold Optimizer ($DPD_{max}=3.5\%$, Acc = 18\%) & Grid Search ($DPD_{max}=53.9\%$, Acc = 45\%)\\ \hline
Equalized Odds Difference (EOD) (\%) & 66\%  & 31.1\% \newline{(0-18 y)} & Threshold Optimizer ($DPD_{max}=3.4\%$, Acc = 18\%) & Correlation Remover ($DPD_{max}=34.1\%$, Acc = 63\%) \\ \hline
\end{tabular}
\caption{Results of Fairlearn + Computer Vision with applying the mitigation algorithms one at a time.}
\label{tab:harm_definitions1}
\end{table}

\begin{table}[h!]
\hspace{-1cm}
\begin{tabular}{|>{\raggedright\arraybackslash}p{3cm}|>{\raggedright\arraybackslash}p{2cm}|>{\raggedright\arraybackslash}p{2cm}|>{\raggedright\arraybackslash}p{3.5cm}|>{\raggedright\arraybackslash}p{3.5cm}|}
\hline
\textbf{Metric} & \textbf{Accuracy of Original Model} & \textbf{Subgroup with Max Bias} &  \textbf{Algorithm with Max Mitigation} & \textbf{Algorithm with Min Mitigation} \\ \hline
Demographic Parity Difference (DPD) (\%) & 66\% &   55\% (30-80y) & Correlation Remover + Exponentiated Gradient ($DPD_{max}=5\%$, Acc = 30.10\%) & Correlation Remover + Grid Search ($DPD_{max}=19.9\%$, Acc = 51.70\%)\\ \hline
Equalized Odds Difference (EOD) (\%) & 66\%  & 31.1\% \newline{(0-18 y)} & Correlation Remover + Exponentiated Gradient ($DPD_{max}=7.8\%$, Acc = 30.1\%) & Correlation Remover + Grid Search ($DPD_{max}=28.1\%$, Acc = 51.7\%) \\ \hline
\end{tabular}
\caption{Results of Fairlearn + Computer Vision with applying the mitigation algorithms in a sequential order.}
\label{tab:harm_definitions2}
\end{table}

\begin{figure}[!h]
\centering
\includegraphics[scale=0.5]{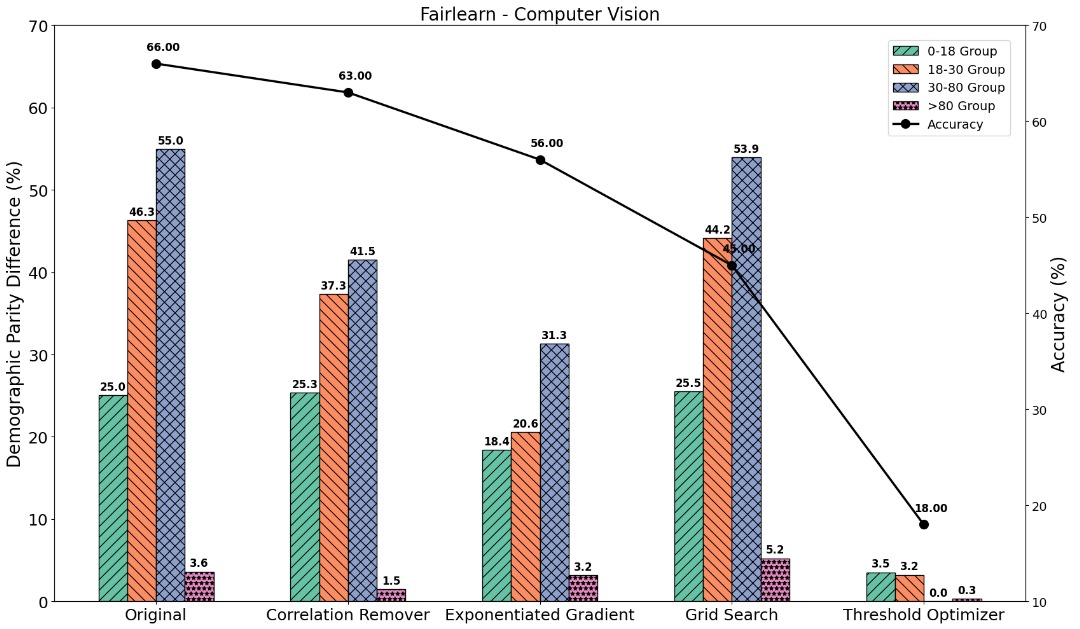}\\
\includegraphics[scale=0.5]{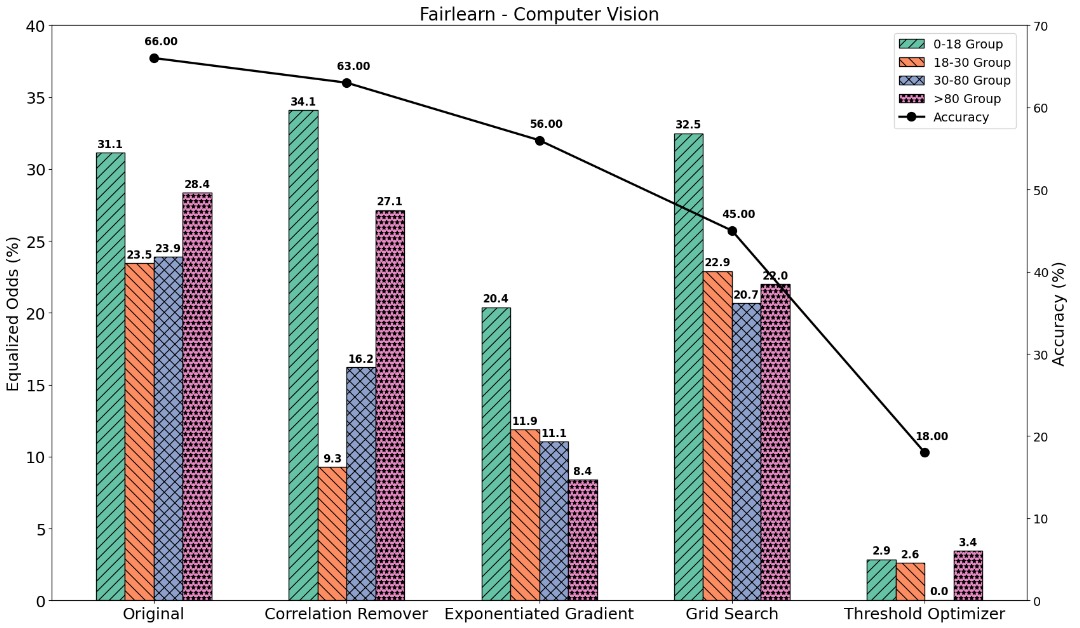}
\caption{The results of applying Fairlearn to the computer vision model with applying the mitigation algorithms one at a time.}
\end{figure}

\begin{figure}[!h]
\centering
\includegraphics[scale=0.265]{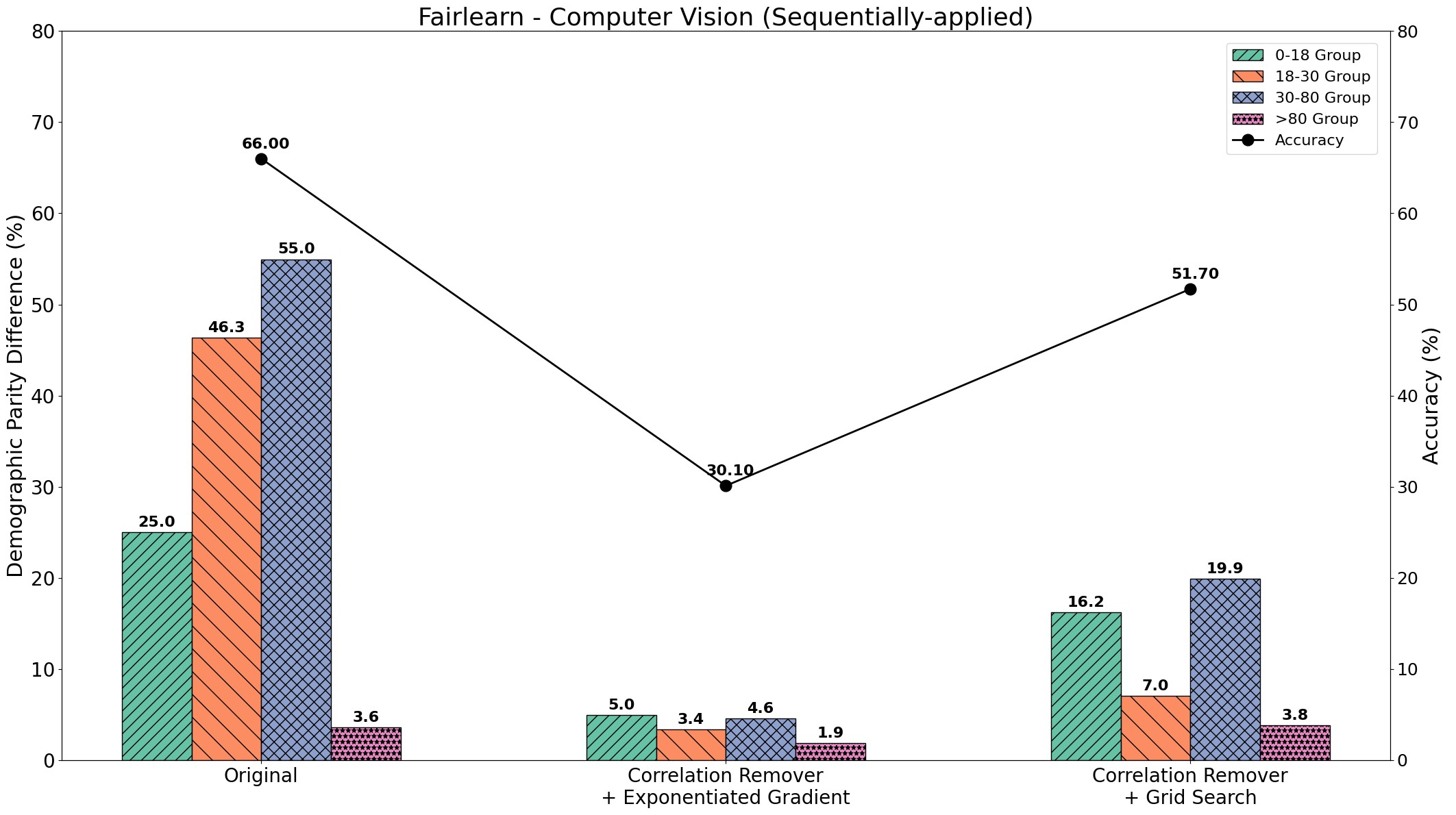}\\
\includegraphics[scale=0.265]{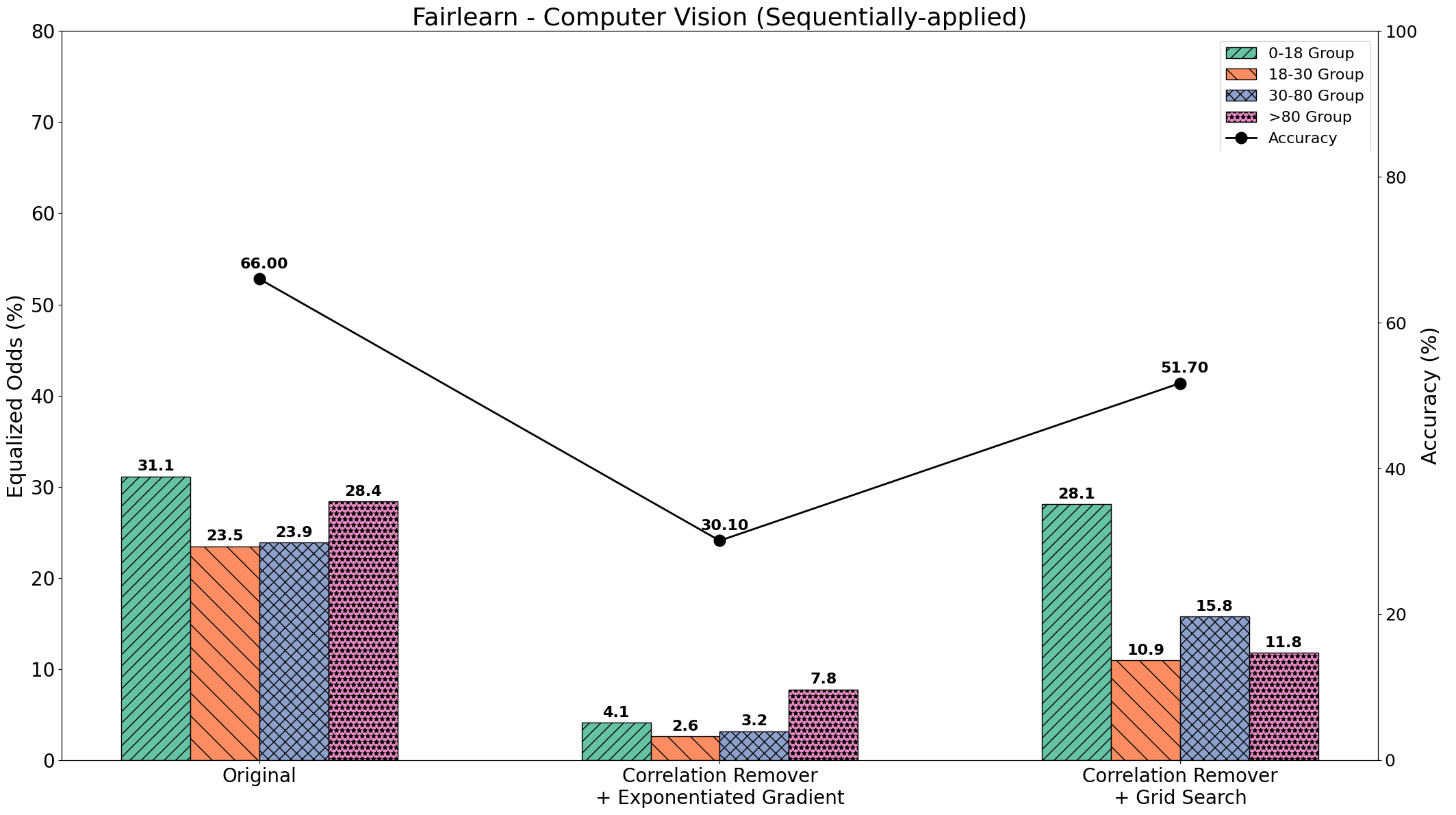}
\caption{The results of applying Fairlearn to the computer vision model with applying the mitigation algorithms in a sequential order.}
\end{figure}

\subsubsection{AIF360 Library}
The AIF360 library is used to analyze bias in the same computer vision problem, taking advantage of its ability to handle multi-class classification directly. In Figure 4, we present various fairness metrics from the AIF360 library, including Statistical Parity Difference, Average Odds Difference, Equal Opportunity Difference, Theil Index, and Generalized Entropy Index. First, we show the values of these metrics for the original model, which has an accuracy of 0.7050, without applying any mitigation algorithms. Then, we demonstrate the impact of applying different mitigation algorithms on both fairness metrics and model accuracy. The algorithms used include Reweighing, Disparate Impact Remover, Learning Fair Representations, Adversarial Debiasing, Equalized Odds Postprocessing, and Reject Option Classification. In Figure 5, we show the effect of applying these mitigation algorithms in sequential order, highlighting their influence on both model bias and performance. The results are discussed below. Five fairness metrics are analyzed:
\begin{enumerate}
\item	Metrics:
\begin{itemize}
\item	Statistical Parity Difference (SPD): Analogous to DPD, measures selection rate differences among sensitive groups.
\item	Average Odds Difference (AOD): Averages the differences in false positive and true positive rates between groups.
\item	Equal Opportunity Difference (EOD): Measures recall disparities between privileged and unprivileged groups.
\item	Generalized Entropy Index: Quantifies prediction randomness across groups to ensure fairness.
\item	Theil Index: Similar to the entropy index but more sensitive to small datasets.
\end{itemize}
\item	Results:\\
The results of the AIF360 metrics with applying the mitigation algorithms one at a time (individually) and in sequential order are shown on Figure 4,5 and the summary of the results are shown in Table (\ref{tab:harm_definitions3}). 
\item	Insights:\\
From the results in Table (\ref{tab:harm_definitions3}), we find that the original model has accuracy of 70.5\% and the highest bias metric (Theil Index) is 58.61\%. 

After applying the mitigation algorithms individually, the highest bias metric (Theil Index) decreased to 47.14\% in the best-case scenario, with accuracy reaching 70.90\% using the Disparate Impact Remover algorithm. In contrast, when applying the mitigation algorithms sequentially, the highest bias metric (Theil Index) dropped to 53.30\%, with accuracy at 62.83\% using the combination of Reweighing + Disparate Impact Remover + Adversarial Debiasing + Reject Option Classification (RW+DIR+AD+ROC). Comparing the two approaches, the individual application reduced bias by 19.57\% while maintaining nearly the same accuracy. Meanwhile, the sequential application reduced bias by 9.06\%, but accuracy decreased by 10.88\%. These results indicate that the individual application outperforms the sequential approach in bias mitigation.
\end{enumerate}

Technical Challenges
\begin{itemize}
\item	Some Fairlearn mitigation algorithms, such as Exponentiated Gradient Reduction, were incompatible with TensorFlow-based models.
\item	The Learning Fair Representation algorithm required excessive computational resources, limiting its utility on local machines.
\end{itemize}

This comparative analysis underscores the trade-offs between bias reduction and model accuracy in fairness-focused ML. While Fairlearn offers simplicity for binary tasks, AIF360 provides a more comprehensive suite for multi-class classification. The study highlights the importance of selecting appropriate method of applying the mitigation algorithms to obtain the highest efficiency by reducing the ML bias and maintaining the model performance.

\begin{table}[h!]
\hspace{-1cm}
\begin{tabular}{|>{\raggedright\arraybackslash}p{2cm}|>{\raggedright\arraybackslash}p{2cm}|>{\raggedright\arraybackslash}p{2cm}|>{\raggedright\arraybackslash}p{4cm}|>{\raggedright\arraybackslash}p{4cm}|}
\hline
\textbf{Mitigation Algorithm} & \textbf{Accuracy of Original Model} & \textbf{Metric with Max Bias} &  \textbf{Algorithm with Max Mitigation} & \textbf{Algorithm with Min Mitigation} \\ \hline
Individual Application & 70.5\% &   58.61\% (Theil Index) & Disparate Impact Remover (Theil Index=47.14\%, Acc = 70.90\%) & Equalized Odds Postprocessing (Theil Index = 63.12\%, Acc = 17.86\%)\\ \hline
sequential Application & 70.5\% &   58.61\% (Theil Index) & RW+DIR+AD+ROC (Theil Index = 53.30\%, Acc = 62.83\%) & RW+DIR+AD+EOP (Theil Index = 54.56\%, Acc = 66.90\%) \\ \hline
\end{tabular}
\caption{Results of AIF360 + Computer Vision with applying the mitigation algorithms one at a time (individual) and in sequential order.}
\label{tab:harm_definitions3}
\end{table}

\begin{figure}[!h]
\centering
\includegraphics[scale=0.3]{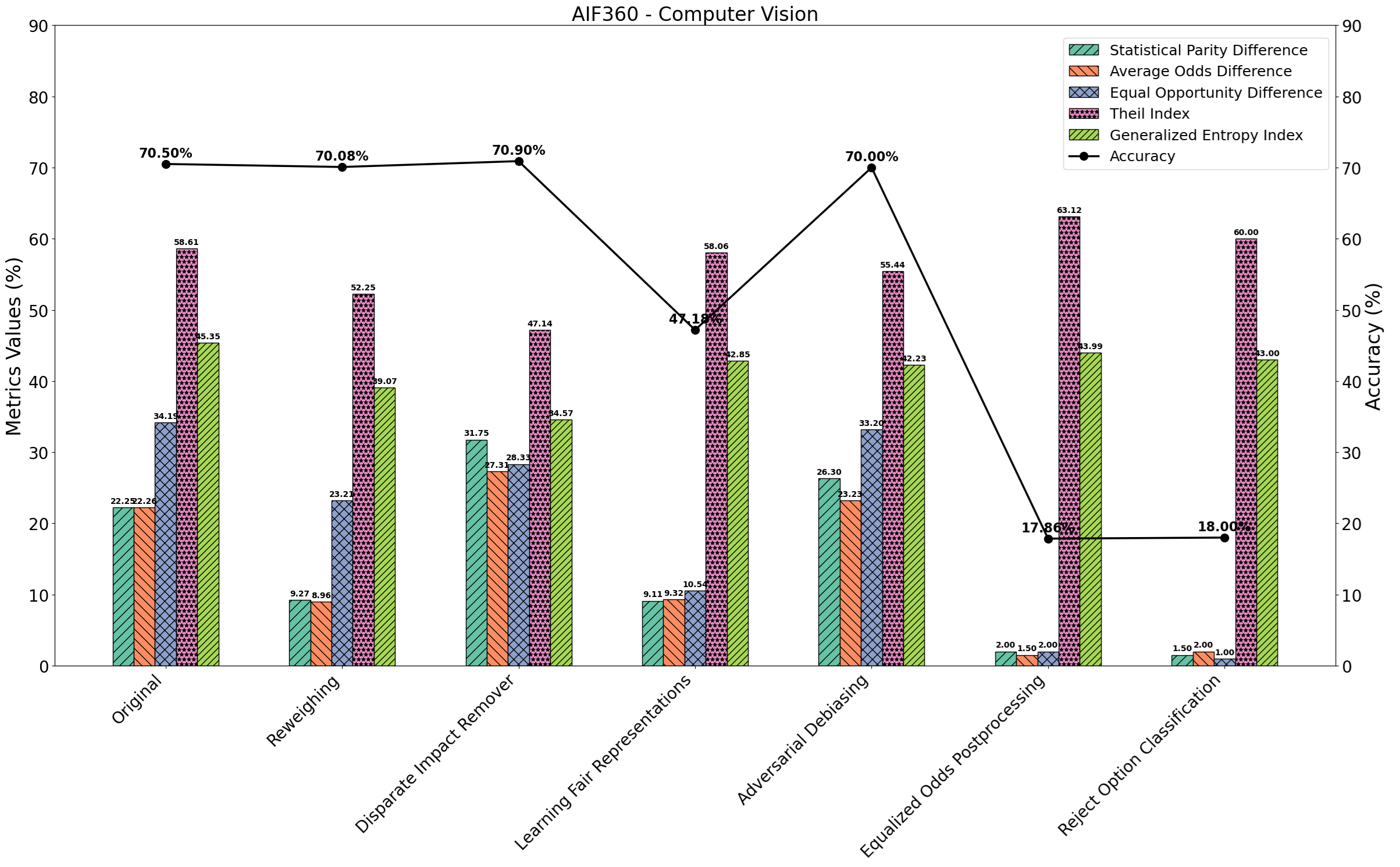}
\caption{The results of applying AIF360 to the computer vision model with applying the mitigation algorithms one at a time.}
\end{figure}

\begin{figure}[!h]
\centering
\includegraphics[scale=0.3]{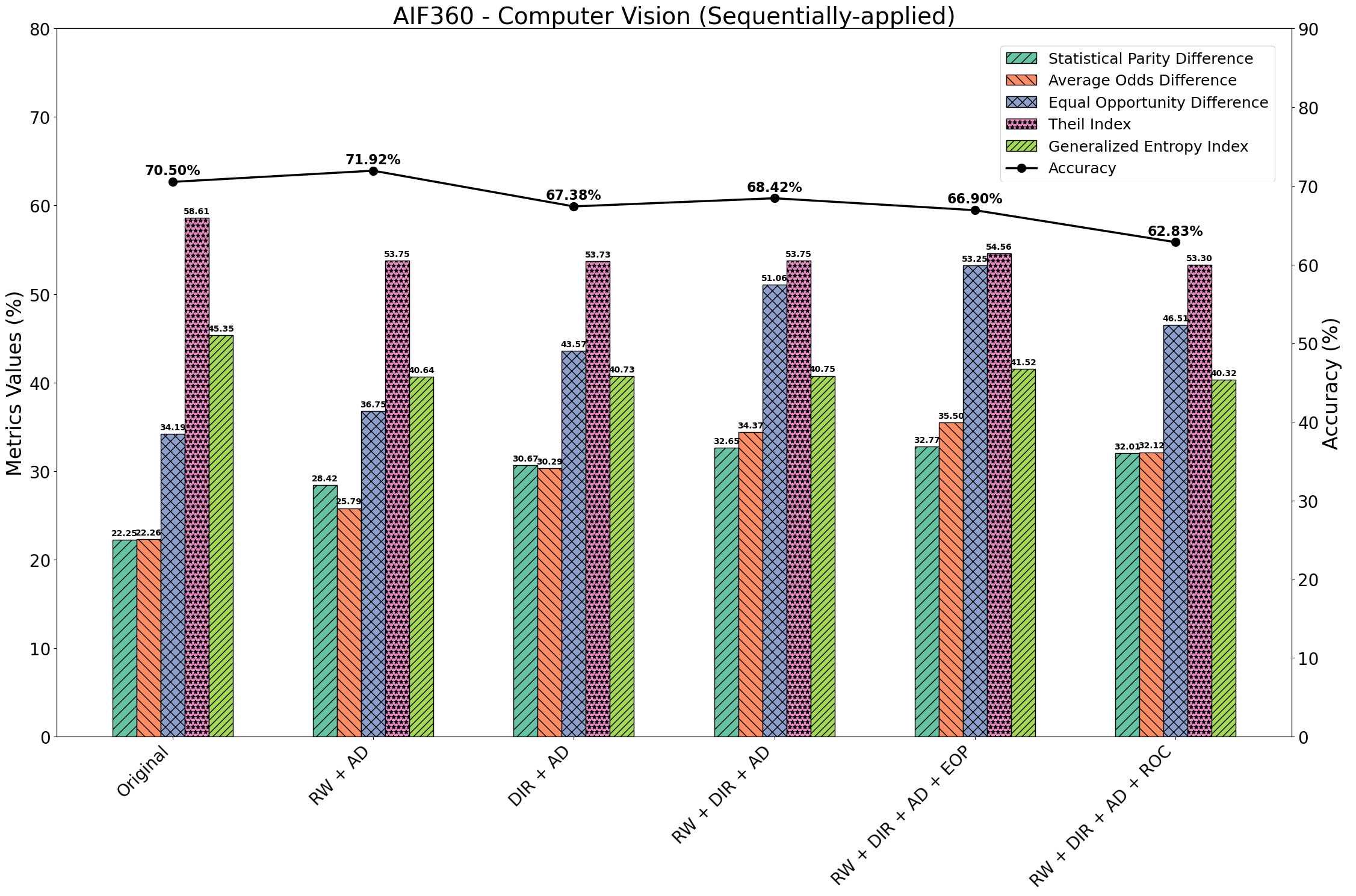}
\caption{The results of applying AIF360 to the computer vision model with applying the mitigation algorithms in a sequential order.}
\end{figure}

\subsection{Bias Mitigation in NLP Models}
\subsubsection{Fairlearn Library}
The Fairlearn library was used to assess fairness in a binary classification task within a natural language processing (NLP) model. Two fairness metrics were evaluated: Demographic Parity Difference (DPD) and Equalized Odds Difference (EOD). Since the model is binary, we present the metrics for each class. In Figure 6, we show the values of these metrics for the original model, which has an accuracy of 0.981. We then demonstrate the effects of applying various mitigation algorithms, including Correlation Remover, Exponentiated Gradient, Grid Search, and Threshold Optimizer, on both fairness and performance metrics. Figure 7 presents the results of applying the mitigation algorithms sequentially. The discussion of the results can be found below.

\begin{enumerate}
\item	Results:\\
The results of the Fairlearn metrics with applying the mitigation algorithms one at a time (individually) and in sequential order are shown on Figure 6,7 and the summary of the results are shown in Table (\ref{tab:harm_definitions4}). 
\item	Insights:\\
From the results in Table (\ref{tab:harm_definitions4}), we observe that the original model has an accuracy of 98.1\% and the highest bias metric (DPD) is 27.4\%.

After applying the mitigation algorithms individually, the highest bias metric (DPD) decreased to 21.4\% with the Exponentiated Gradient algorithm, while the accuracy dropped to 95.4\%. On the other hand, when the mitigation algorithms were applied in a sequential order, the highest bias metric (DPD) decreased to 17.9\%, with an accuracy of 92.7\% using the combination of Correlation Remover and Exponentiated Gradient.

Comparing the individual and sequential applications of the algorithms, we find that the individual application reduces bias by 21.89\%, but the model’s accuracy decreases by 2.75\%. In contrast, the sequential application reduces bias by 34.67\%, but the model’s accuracy decreases by 5.5\%. These results show that the sequential application is more effective in reducing bias, while maintaining model performance at a similar level.
\end{enumerate}

\begin{table}[h!]
\hspace{-1cm}
\begin{tabular}{|>{\raggedright\arraybackslash}p{2cm}|>{\raggedright\arraybackslash}p{2cm}|>{\raggedright\arraybackslash}p{2cm}|>{\raggedright\arraybackslash}p{4cm}|>{\raggedright\arraybackslash}p{4cm}|}
\hline
\textbf{Mitigation Algorithm} & \textbf{Accuracy of Original Model} & \textbf{Metric with Max Bias} &  \textbf{Algorithm with Max Mitigation} & \textbf{Algorithm with Min Mitigation} \\ \hline
Individual Application & 98.1\% &   27.4\% (DPD) & Exponentiated Gradient (DPD = 21.4\%, Acc = 95.4\%) & Grid Search (EOD = 58.3\%, Acc = 90\%)\\ \hline
sequential Application & 98.1\% &   27.4\% (DPD) & Correlation Remover + Exponentiated Gradient (DPD = 17.9\%, Acc = 92.7\%) & Correlation Remover + Grid Search (EOD = 41.3\%, Acc = 93.6\%) \\ \hline
\end{tabular}
\caption{Results of Fairlearn + NLP with applying the mitigation algorithms one at a time (individual) and in sequential order.}
\label{tab:harm_definitions4}
\end{table}

\begin{figure}[!h]
\centering
\includegraphics[scale=0.55]{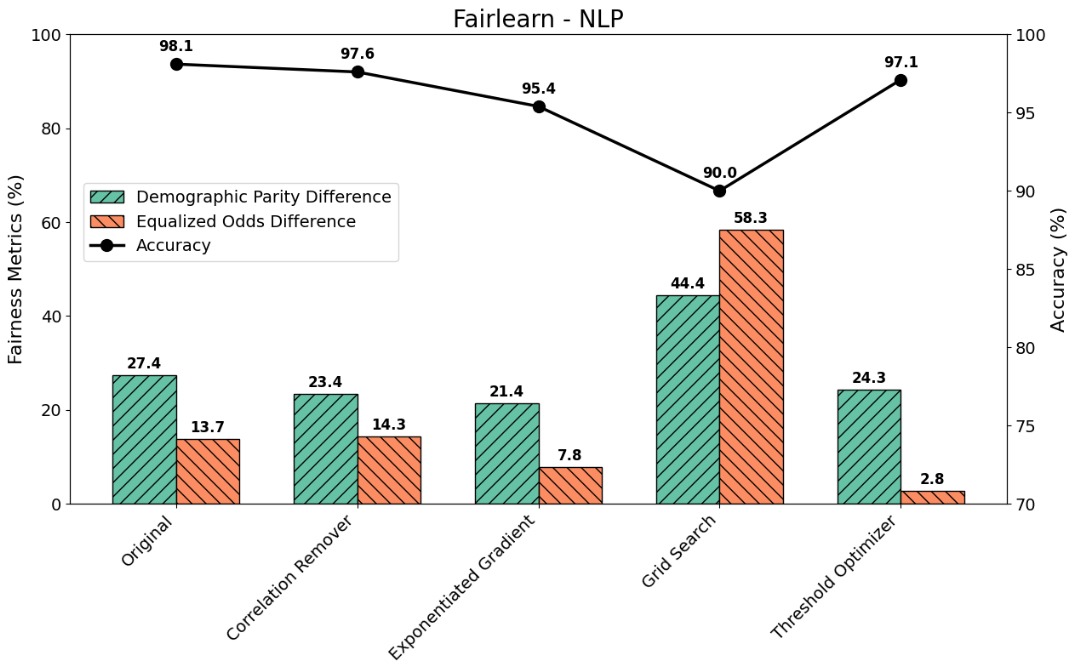}
\caption{The results of applying Fairlearn to the NLP model with applying the mitigation algorithms one at a time.}
\end{figure}

\begin{figure}[!h]
\centering
\includegraphics[scale=0.55]{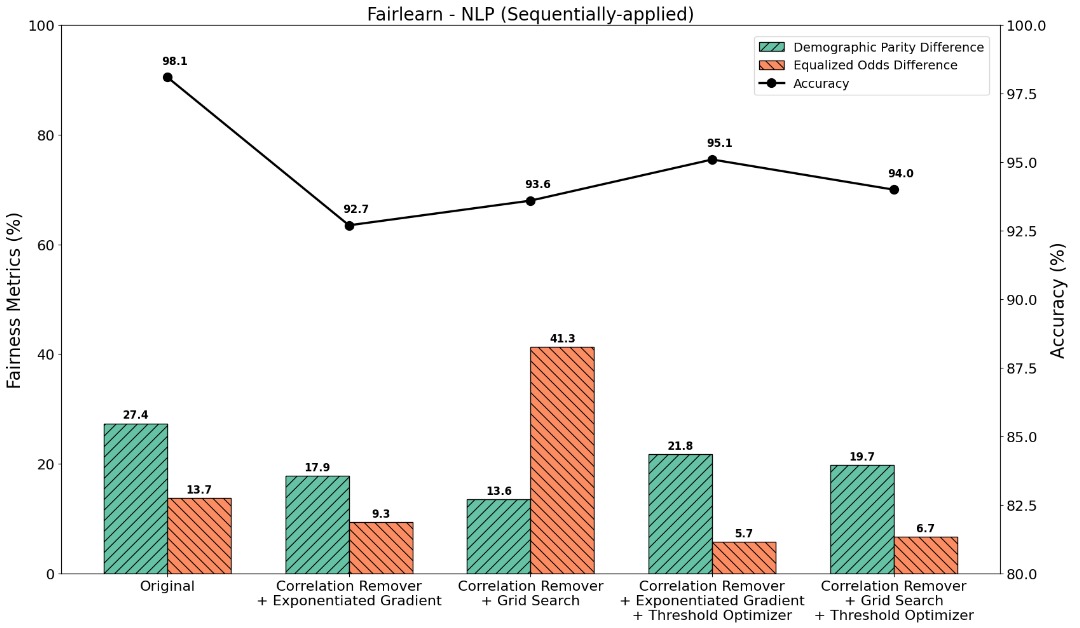}
\caption{The results of applying Fairlearn to the NLP model with applying the mitigation algorithms in a sequential order.}
\end{figure}

\subsubsection{AIF360 Library}

The AIF360 library was also applied to evaluate fairness in the same NLP model, using the same fairness metrics for assessment. Figure 8 presents various fairness metrics available in the AIF360 library, such as statistical parity difference, average odds difference, equal opportunity difference, Theil index, and generalized entropy index. Initially, we display the values of these metrics for the original model, which achieves an accuracy of 98.14\%, without applying any mitigation algorithms. Following this, we analyze the impact of different mitigation algorithms on both fairness metrics and model accuracy. The mitigation techniques used include reweighing, disparate impact remover, learning fair representations, adversarial debiasing, equalized odds postprocessing, and reject option classification. Figure 9 shows the effect of applying these mitigation algorithms in a sequential order. The summarized results, highlighting the analysis of these five fairness metrics, are shown in Table \ref{tab:harm_definitions5}.

\begin{enumerate}
\item	Results:\\
The results of the AIF360 metrics with applying the mitigation algorithms one at a time (individually) and in sequential order are shown on Figure 8,9 and the summary of the results are shown in Table (\ref{tab:harm_definitions5}). 
\item	Insights:\\
From the results in Table (\ref{tab:harm_definitions5}), we find that the original model has accuracy of 98.14\% and the highest bias metric (SPD) is 19.80\%. 

After applying the mitigation algorithms individually, the highest bias metric (SPD) decreased to 19.71\%, with the model's accuracy reaching 97.23\% using the Reweighing algorithm. In contrast, when the mitigation algorithms were applied in a sequential order, the highest bias metric (SPD) remained unchanged, while the accuracy slightly dropped to 97.32\% with the combination of Reweighing + Disparate Impact Remover + Adversarial Debiasing + Equalized Odds Postprocessing (RW+DIR+AD+EOP). A comparison of the individual and sequential application of algorithms reveals that both methods resulted in minimal change to the model's bias or performance.
\end{enumerate}

\begin{table}[h!]
\hspace{-1cm}
\begin{tabular}{|>{\raggedright\arraybackslash}p{2cm}|>{\raggedright\arraybackslash}p{2cm}|>{\raggedright\arraybackslash}p{2cm}|>{\raggedright\arraybackslash}p{4cm}|>{\raggedright\arraybackslash}p{4cm}|}
\hline
\textbf{Mitigation Algorithm} & \textbf{Accuracy of Original Model} & \textbf{Metric with Max Bias} &  \textbf{Algorithm with Max Mitigation} & \textbf{Algorithm with Min Mitigation} \\ \hline
Individual Application & 98.14\% &   19.79\% (SPD) & Reweighing (SPD = 19.71\%, Acc = 97.23\%) & Learning Fair Representations (SPD = 25.81\%, Acc = 72.89\%)\\ \hline
sequential Application & 98.14\% &   19.80\% (SPD) & RW+DIR+AD+EOP (SPD = 19.80\%, Acc = 97.32\%) & RW+DIR+AD+ROC (SPD = 20.16\%, Acc = 97.46\%) \\ \hline
\end{tabular}
\caption{Results of AIF360 + NLP with applying the mitigation algorithms one at a time (individual) and in sequential order.}
\label{tab:harm_definitions5}
\end{table}

\begin{figure}[!h]
\centering
\includegraphics[scale=0.27]{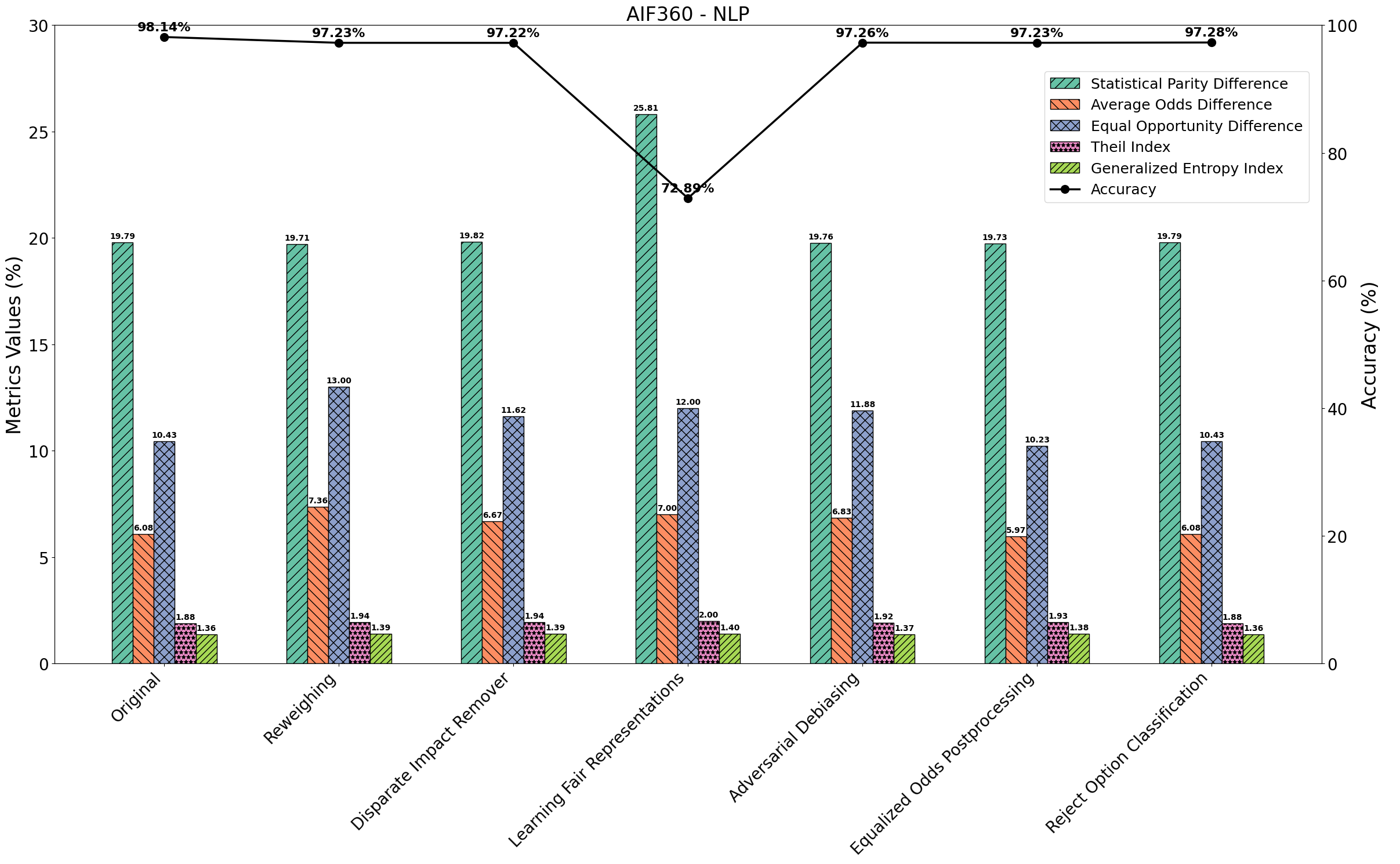}
\caption{The results of applying AIF360 to the NLP model with applying the mitigation algorithms one at a time.}
\end{figure}

\begin{figure}[!h]
\centering
\includegraphics[scale=0.27]{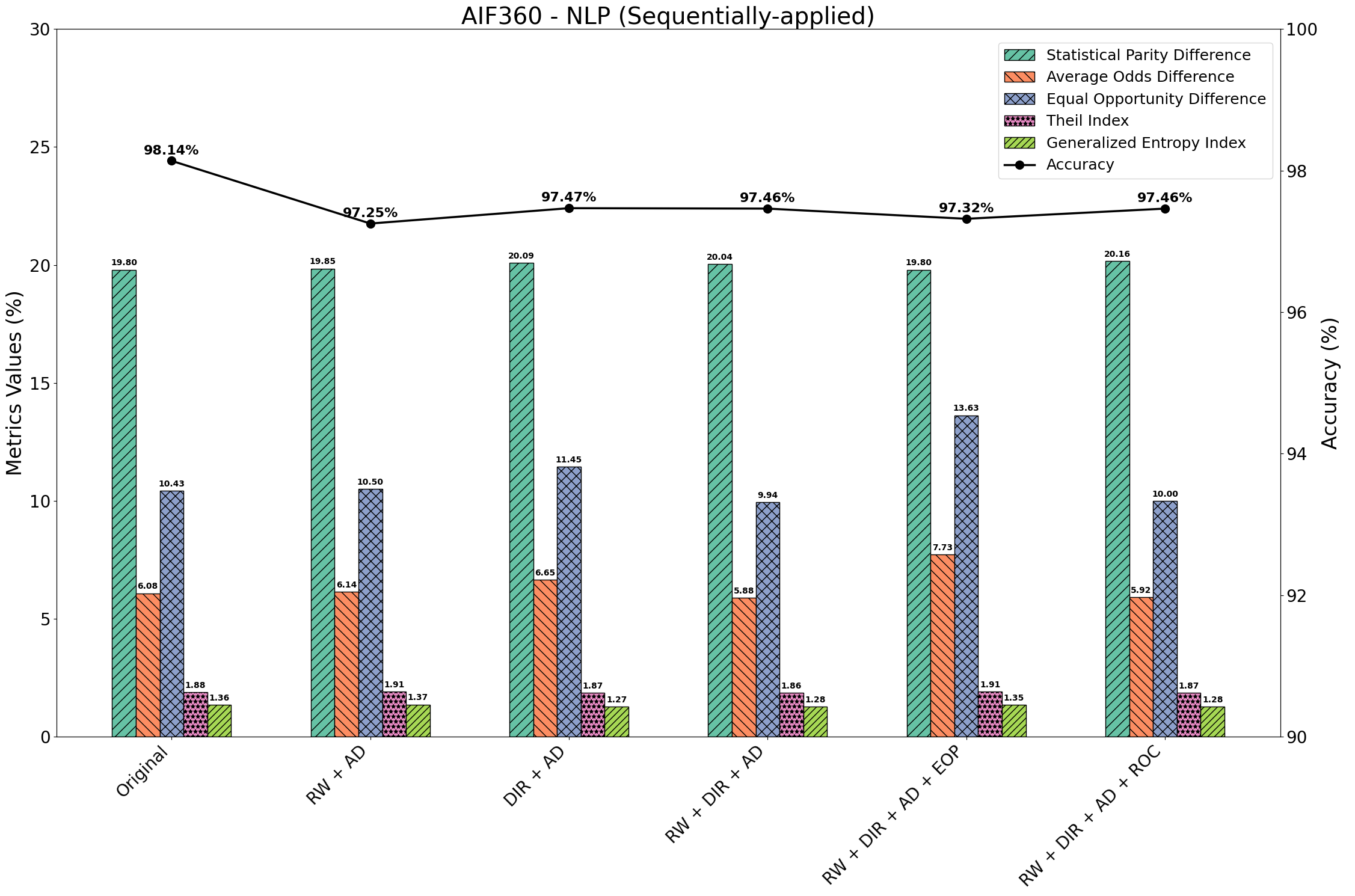}
\caption{The results of applying AIF360 to the NLP model with applying the mitigation algorithms in a sequential order.}
\end{figure}

\section{Future Work}

One of the key challenges in applying fairness interventions to generative AI models is the complexity of their architectures and outputs [42]. Unlike traditional models, generative models produce unstructured data such as text, images, and other media, making it more difficult to detect and mitigate bias [43]. To address this, fairness metrics specifically designed for generative models—such as distribution-level evaluations and fairness-aware assessments of generation quality—need to be developed. Additionally, the multi-modal nature of these models requires advancements in fairness algorithms to handle diverse data types effectively. A major obstacle remains the lack of benchmark datasets explicitly designed for fairness evaluation in generative AI. Creating such datasets and integrating them into fairness libraries would enable more robust testing and comparison of mitigation strategies.

The techniques discussed in this paper can be extended to generative AI models, including Generative Adversarial Networks (GANs), Variational Autoencoders (VAEs), and transformer-based models like GPT and DALL-E. Applying sequential mitigation algorithms across the three stages of the ML lifecycle would help identify and reduce biases in both their training processes and outputs. This approach aligns with recent findings that mitigation strategies can be effective when applied holistically across the ML lifecycle [44]. It holds significant potential in areas like content creation, data augmentation, and conversational AI, where fairness and ethical considerations are critical [45]. Another important extension is exploring how sequential fairness interventions impact the creative capacities of generative models, ensuring diversity and representativeness are preserved without compromising quality [46].

Extending fairness interventions to generative AI models has a broader impact on promoting equity and inclusivity in AI-generated content. Generative models are increasingly used in domains such as art, education, entertainment, and media, where biased outputs could perpetuate harmful stereotypes or exclusion [47]. Addressing fairness in these models ensures that they contribute to a more inclusive digital ecosystem [48]. This research will provide actionable insights and tools for practitioners to incorporate fairness into generative AI systems, fostering trust and accountability in AI deployment [48]. Ultimately, applying fairness principles to generative models will transform how these systems are designed, evaluated, and utilized, ensuring that their outputs align with ethical standards and societal values.

Explainability in generative AI systems remains a significant challenge, making it difficult to understand how biases are introduced and propagated. To address this, integrating Explainable AI (XAI) techniques is essential. These techniques can reveal the inner workings of generative models and enable targeted fairness interventions at key points [49]. Furthermore, the absence of benchmark datasets for evaluating fairness in multi-modal generative models underscores the need for diverse, high-quality datasets that capture potential biases and facilitate rigorous testing and validation of fairness strategies [43]. XAI can also enhance model clarity, offering stakeholders insights into how fairness interventions influence outcomes [50]. Future research could explore fairness-aware generation in multi-modal conversational agents, ensuring equitable interactions across languages, cultures, and user demographics [46]. By integrating XAI into generative models, users can understand and challenge biases, promoting accountability and ethical AI practices. Extending fairness interventions to multi-modal and generative systems supports global efforts to develop AI that reflects societal values and ethical standards, contributing to a more equitable digital ecosystem.

\section{Conclusion}

In this work, we studied the fairness of the machine learning models with using unstructured data. We used the fairness libraries Fairlearn by Microsoft and AIF360 by IBM, and used Kaggle datasets including images with a computer vision model and text with natural language processing model. This research provided practical insights of a comparative study between  applying the mitigation algorithms one at a time (individually) in the stages of the ML lifecycle and in more than a stage in a sequential order in both models using both libraries. The study highlighted the role of biased predictions in reinforcing systemic inequalities and underscores the need for evaluating and improving fairness in ML workflows. 

From the results of applying the fairlearn and AIF360 mitigation algorithms to the computer vision model, we found that fairlearn reduced the bias in the model substantially better than AIF360, but at the cost of dropping the model performance. By applying the mitigation algorithms in a sequential order, fairlearn was able to maintain the model performance by about 67.22\% over that of applying the algorithms one at a time with reducing the bias almost to the same level. On the other hand, the individual application of the algorithms in AIF360 library showed better performance over the sequential order application on the level of the model fairness and performance. 

In the case of the NLP model, we found that fairlearn reduced the bias in the model  better than AIF360 which nearly did not change neither the fairness nor the performance of the model. By applying the mitigation algorithms in a sequential order, fairlearn was able to slightly reduce the bias better than applying individually by a factor of 16.35\%.



Our findings indicated that applying mitigation algorithms in a sequential order can produce favorable results compared to using a single algorithm. This approach successfully reduced bias while preserving model performance in some cases, as reflected by minimal changes in accuracy. By showing that a sequential strategy can balance fairness and performance. Our work introduced a novel and practical framework for research and industrial efforts in mitigating ML bias.

\section*{Acknowledgement}

We would like to thank Yale Center for Research Computing for supporting A.K. through the CAREERS project administered by the PSU Institute for Computational and Data Sciences (ICDS) under the National Science Foundation Award with No. 2018873.

\section*{REFERENCES}
\noindent
[1] Amitabha Mukerjee, Rita Biswas, Kalyanmoy Deb, and Amrit P. Mathur. 2002. Multi–objective evolutionary algorithms for the risk–return trade–off in bank loan management. Int. Trans. Oper. Res. 9, 5 (2002), 583–597.

\noindent
[2] Miranda Bogen and Aaron Rieke. 2018. HelpWanted: An Examination of Hiring Algorithms, Equity and Bias. Technical Report. Upturn.

\noindent
[3] Lee Cohen, Zachary C. Lipton, and Yishay Mansour. 2019. Efficient candidate screening under multiple tests and implications for fairness. arXiv:cs.LG/1905.11361 (2019).

\noindent
[4] Shai Danziger, Jonathan Levav, and Liora Avnaim-Pesso. 2011. Extraneous factors in judicial decisions. Proc. Nat. Acad. Sci. 108, 17 (2011), 6889–6892.

\noindent
[5] Anne O’Keeffe and Michael McCarthy. 2010. The Routledge Handbook of Corpus Linguistics. Routledge.

\noindent
[6] Julia Angwin, Jeff Larson, Surya Mattu, and Lauren Kirchner. 2019. Machine bias: There’s software used across the country to predict future criminals. and it’s biased against blacks. https://www.propublica.org/article/machine-biasrisk-assessments-in-criminal-sentencing.

\noindent
[7] Cathy O’Neil. 2016.Weapons of Math Destruction: How Big Data Increases Inequality and Threatens Democracy. Crown Publishing Group, New York, NY.

\noindent
[8] M. A. Madaio, L. Stark, J. Wortman Vaughan, and H. Wallach. Co-Designing Checklists to Understand Organizational Challenges and Opportunities around Fairness in AI. Chi 2020, pages 1–14, 2020.

\noindent
[9] J. Buolamwini and T. Gebru. Gender shades: Intersectional accuracy disparities in commercial gender classification. In Conference on fairness, accountability and transparency, pages 77–91, 2018.

\noindent
[10] A. Guo, E. Kamar, J. W. Vaughan, H. M. Wallach, and M. R. Morris. Toward fairness in AI for people with disabilities: A research roadmap. CoRR, abs/1907.02227, 2019. URL http://arxiv.org/abs/1907.02227.

\noindent
[11] C. Rudin, C. Wang, and B. Coker. The age of secrecy and unfairness in recidivism prediction. pages 1–46, 2018. URL http://arxiv.org/abs/1811.00731.

\noindent
[12] Tolga Bolukbasi, Kai-Wei Chang, James Zou, Venkatesh Saligrama, Adam Kalai.
“Man is to Computer Programmer as Woman is to Homemaker? Debiasing Word Embeddings”, arXiv:1607.06520

\noindent
[13] Denis Rystsov. “CASPaxos: Replicated State Machines without logs”. arXiv:1802.07000

\noindent
[14] J.P.C.Greenlees. “The Balmer spectrum of rational equivariant cohomology theories”. arXiv:1706.07868

\noindent
[15] Ahmed Rashed, Abdelkrim Kallich, Mohamed Eltayeb. "Analyzing Fairness of Classification Machine Learning Model with Structured Dataset". arXiv:2412.09896 [cs.LG].

\noindent
[16] Computer Vision Dataset: https://www.kaggle.com/datasets/jangedoo/utkface-new \\
         NLP Dataset: https://www.kaggle.com/datasets/prasad22/ca-independent-medical-review

\noindent
[17] Shahriar S, Allana S, Hazratifard SM, Dara R. "A survey of privacy risks and mitigation strategies in the Artificial Intelligence life cycle". IEEE Access. 2023 Jun 19;11:61829-54.

\noindent
[18] Mehrabi, N., Morstatter, F., Saxena, N., Lerman, K., \& Galstyan, A. (2021). A Survey on Bias and Fairness in Machine Learning. ACM Computing Surveys, 54(6), 1-35.

\noindent
[19] Suresh, H., \& Guttag, J. V. (2021). A Framework for Understanding Unintended Consequences of Machine Learning. Communications of the ACM, 64(8), 62-71.

\noindent
[20] Saber Malekmohammadi, Afaf Taik, Golnoosh Farnadi. "Mitigating Disparate Impact of Differential Privacy in Federated Learning through Robust Clustering". arXiv:2405.19272 [cs.LG]

\noindent
[21] Andrew Bell, Joao Fonseca, Carlo Abrate, Francesco Bonchi, Julia Stoyanovich. "Fairness in Algorithmic Recourse Through the Lens of Substantive Equality of Opportunity". arXiv:2401.16088 [cs.LG].

\noindent
[22] Shizhou Xu, Thomas Strohmer. "On the (In)Compatibility between Group Fairness and Individual Fairness". arXiv:2401.07174 [math.ST]

\noindent
[23] Binns, R. (2018). Fairness in Machine Learning: Lessons from Political Philosophy. Proceedings of the 2018 Conference on Fairness, Accountability, and Transparency (FAT).

\noindent
[24] Di Wu. "The effects of data preprocessing on probability of default model fairness". arXiv:2408.15452 [econ.EM]

\noindent
[25] Zhang, B. H., Lemoine, B., \& Mitchell, M. (2018). Mitigating Unwanted Biases with Adversarial Learning. Proceedings of the 2018 AAAI/ACM Conference on AI, Ethics, and Society (AIES).

\noindent
[26] Chouldechova, A. (2017). Fair Prediction with Disparate Impact: A Study of Bias in Recidivism Prediction Instruments. Big Data, 5(2), 153-163.

\noindent
[27] Angwin, J., Larson, J., Mattu, S., \& Kirchner, L. (2016). Machine Bias. ProPublica.

\noindent
[28] Xu, D., Yuan, S., Zhang, L., \& Wu, X. (2020). FairGAN: Fairness-aware Generative Adversarial Networks. Proceedings of the 2020 International Joint Conference on Artificial Intelligence (IJCAI).

\noindent
[29] Navid Nayyem, Abdullah Rakin, Longwei Wang. "Bridging Interpretability and Robustness Using LIME-Guided Model Refinement". arXiv:2412.18952 [cs.LG]

\noindent
[30] Lundberg, S. M., \& Lee, S.-I. (2017). A Unified Approach to Interpreting Model Predictions. Proceedings of the 31st Conference on Neural Information Processing Systems (NeurIPS).

\noindent
[31] Buolamwini, J., \& Gebru, T. (2018). Gender Shades: Intersectional Accuracy Disparities in Commercial Gender Classification. Proceedings of the 2018 Conference on Fairness, Accountability, and Transparency (FAT).

\noindent
[32] Kearns, M., Neel, S., Roth, A., \& Wu, Z. S. (2018). Preventing Fairness Gerrymandering: Auditing and Learning for Subgroup Fairness. Proceedings of the 35th International Conference on Machine Learning (ICML).

\noindent
[33] Barocas, S., Hardt, M., \& Narayanan, A. (2019). Fairness and Accountability in Machine Learning. Proceedings of the 2019 Conference on Fairness, Accountability, and Transparency.

\noindent
[34] Angwin, J., Larson, J., Mattu, S., \& Kirchner, L. (2016). Machine Bias: There's Software Used Across the Country to Predict Future Criminals. And it's Biased Against Blacks. ProPublica.

\noindent
[35] Fairlearn by Microsoft. (n.d.). Retrieved from https://fairlearn.org/ 

\noindent
[36] AIF360 by IBM. (n.d.). Retrieved from https://aif360.mybluemix.net/

\noindent
[37] What-If Tool by Google. (n.d.). Retrieved from https://github.com/google/tf-what-if 

\noindent
[38] Github of the project: https://github.com/mohammad2012191/Fairness-in-Machine-Learning-Identifying-and-Mitigation-of-Bias/ 

\noindent
[39] Moritz Hardt, Eric Price, and Nati Srebro. Equality of opportunity in supervised learning. In NeurIPS, 3315–3323. 2016.\\ URL: https://proceedings.neurips.cc/paper/2016/hash/9d2682367c3935defcb1f9e247a97c0d-Abstract.html 

\noindent
[40] Hilde Weerts, Lambèr Royakkers, and Mykola Pechenizkiy. Does the end justify the means? on the moral justification of fairness-aware machine learning. arXiv preprint arXiv:2202.08536, 2022.

\noindent
[41] Brent Mittelstadt, Sandra Wachter, and Chris Russell. The unfairness of fair machine learning: levelling down and strict egalitarianism by default. arXiv preprint arXiv:2302.02404, 2023.

\noindent
[42] Barocas, S., Hardt, M., \& Narayanan, A. (2017). Fairness and machine learning. Online resource

\noindent
[43] Bellamy, R. K. E., et al. (2019). AI Fairness 360: An extensible toolkit for detecting, understanding, and mitigating unwanted algorithmic bias. IBM Journal of Research and Development. DOI:10.1147/JRD.2019.2942287

\noindent
[44] Bolukbasi, T., et al. (2016). Man is to computer programmer as woman is to homemaker? Debiasing word embeddings. NeurIPS 2016. DOI:10.48550/arXiv.1607.06520

\noindent
[45] Friedler, S. A., et al. (2019). A comparative study of fairness-enhancing interventions in machine learning. Proceedings of FAT 2019. DOI:10.48550/arXiv.1802.04422

\noindent
[46] Mehrabi, N., et al. (2021). A survey on bias and fairness in machine learning. ACM Computing Surveys. DOI:10.1145/3457607

\noindent
[47] Rajkomar, A., et al. (2018). Ensuring fairness in machine learning to advance health equity. Annals of Internal Medicine. DOI:10.7326/M18-1990

\noindent
[48] Xu, D., et al. (2018). FairGAN: Fairness-aware generative adversarial networks. 2018 IEEE Big Data. DOI:10.1109/BigData.2018.8622626

\noindent
[49] Arrieta, A. B., et al. (2020). Explainable artificial intelligence (XAI): Concepts, taxonomies, opportunities, and challenges toward responsible AI. Information Fusion. DOI:10.1016/j.inffus.2019.12.012

\noindent
[50] Doshi-Velez, F., \& Kim, B. (2017). Towards a rigorous science of interpretable machine learning. arXiv preprint. DOI:10.48550/arXiv.1702.08608

\end{document}